\documentclass{article}

\usepackage{microtype}
\usepackage{graphicx}
\usepackage{subfigure}
\usepackage{amsmath}
\usepackage{amssymb}
\usepackage{bm}
\usepackage{booktabs} 
\usepackage{mathrsfs}
\usepackage{algorithm}
\usepackage{algorithmic}

\usepackage{hyperref}

\usepackage[accepted]{icml2020}

\icmltitlerunning{Self-PU: Self Boosted and Calibrated Positive-Unlabeled Training}

\begin{document}

\twocolumn[
\icmltitle{Self-PU: Self Boosted and Calibrated Positive-Unlabeled Training}



\icmlsetsymbol{equal}{*}

\begin{icmlauthorlist}
\icmlauthor{Xuxi Chen*}{ustc}
\icmlauthor{Wuyang Chen*}{tamu}
\icmlauthor{Tianlong Chen}{tamu}
\icmlauthor{Ye Yuan}{tamu}
\icmlauthor{Chen Gong}{njust}
\icmlauthor{Kewei Chen}{banner}
\icmlauthor{Zhangyang Wang}{tamu}
\end{icmlauthorlist}

\icmlaffiliation{tamu}{Texas A\&M University}
\icmlaffiliation{ustc}{University of Science and Technology of China}
\icmlaffiliation{banner}{Banner Alzheimer's Institute}
\icmlaffiliation{njust}{Nanjing University of Science and Technology}

\icmlcorrespondingauthor{Zhangyang Wang}{atlaswang@tamu.edu}

\icmlkeywords{Machine Learning, ICML}

\vskip 0.3in
]



\printAffiliationsAndNotice{\icmlEqualContribution The work was done when Xuxi Chen was mentored by Zhangyang Wang.} 

\begin{abstract}
Many real-world applications have to tackle the Positive-Unlabeled (PU) learning problem, \textit{i.e.}, learning binary classifiers from a large amount of unlabeled data
and a few labeled positive examples.
While current state-of-the-art methods employ importance reweighting to design various risk estimators, they ignored the learning capability of the model itself, which could have provided reliable supervision.
This motivates us to propose a novel \textbf{Self-PU} learning framework, which seamlessly integrates PU learning and self-training.
Self-PU highlights three ``self''-oriented building blocks: a \textit{self-paced} training algorithm that adaptively discovers and augments confident positive/negative examples as the training proceeds; a \textit{self-calibrated} instance-aware loss; and a \textit{self-distillation} scheme that introduces teacher-students learning as an effective regularization for PU learning. We demonstrate the state-of-the-art performance of Self-PU on common PU learning benchmarks (MNIST and CIFAR-10), which compare favorably against the latest competitors. Moreover, we study a real-world application of PU learning, \textit{i.e.}, classifying brain images of Alzheimer's Disease. Self-PU obtains significantly improved results on the renowned Alzheimer's Disease Neuroimaging Initiative (ADNI) database over existing methods. The code is publicly available at: \url{https://github.com/TAMU-VITA/Self-PU}.
\end{abstract}

\section{Introduction}
For standard supervised learning of binary classifiers, both positive and negative classes need to be collected for training purposes. However, this is not always a realistic setting in many applications, where one certain class of data could be difficult to be collected or annotated. For example, in chronic disease diagnosis, while we might safely consider a diagnosed patient to be ``positive'', the much larger population of ``undiagnosed'' individuals are practically mixed with both ``positive'' (patient) and ``negative'' (healthy) examples, since people might be undergoing the disease's incubation period \cite{armenian1974distribution} or might just have not seen doctors. Roughly labeling the ``undiagnosed'' examples all as negative will hence lead to biased classifiers that inevitably underestimate the risk of chronic disease.

Given those practical demands, Positive-Unlabeled (PU) Learning has been increasingly studied in recent years, where a binary classifier is to be learned from a part of positive examples, plus an unlabeled sample pool of mixed and unspecified positive and negative examples. Because of this weak supervision, PU learning is more challenging than standard supervised or semi-supervised classification problems.
Early works tried to identify reliable negative examples from the unlabeled data by hand-crafted heuristics or standard semi-supervised learning methods \cite{liu2002partially,li2003learning}.
Recently, importance reweighting methods such as unbiased PU (uPU) \cite{du2014analysis,du2015convex} and non-negative PU (nnPU) \cite{kiryo2017positive} treat unlabeled data as weighted negative ones. 

Despite these successes, self-supervision via auxiliary or surrogate tasks was never considered, which could potentially supply another means of reliable supervision. This motivates us to explore the learning capability of the model itself.
Our proposed \textbf{Self-PU} learning framework exploits three aspects of such ``self-boosts'': (a) we design a self-paced training strategy to progressively select unlabeled examples and update the ``trust'' set of confident examples; (b) we explore a fine-grained calibration of the functions for unconfident examples in a meta-learning fashion; and (c) we construct a collaborative self-supervision between teacher and student models, and enforce their consistency as a new regularization, against the weak supervision in PU learning.
Our main contributions are outlined as follows:
\begin{itemize}
    \vspace{-1em}\item A novel \textbf{self-paced} learning pipeline is first introduced to adaptively mine confident examples from unlabeled data, that will be labeled into trusted positive/negative classes. A hybrid loss is applied to both the augmented ``labeled'' examples and remaining unlabeled data for supervision. The procedure is repeated progressively, with more unlabeled examples selected each time.
    \vspace{-0.5em}
    \item A \textbf{self-calibration} strategy is leveraged to further explore the fine-grained treatment of loss functions over unconfident examples, in a meta-learning fashion.
    \vspace{-0.5em}
    \item A \textbf{self-distillation} scheme is designed via the collaborative training between several teacher networks and student networks, providing a consistency regularization as another fold of self-supervision.
    \vspace{-0.5em}
    \item In addition to standard benchmarks (MNIST, CIFAR-10), a \textbf{new real-world testbed} of PU learning, \textit{i.e.}, Alzheimer’s Disease neuroimage classification, is evaluated for the first time. On the Alzheimer’s Disease Neuroimaging Initiative (ADNI) database, Self-PU achieves superior results over existing solutions.
\end{itemize}

\section{Related Work}
\subsection{PU Learning}\label{sec:pu}
Let $X\in \mathop{\mathbb{R}}^d$ and $Y\in\{+1, -1\}$ ($d \in \mathop{\mathbb{N}}$) be the input and output random variables. In PU learning, the training dataset $D$ is composed of a positive set $D_P$ and an unlabeled set $D_U$, where we have $D = D_P \cup D_U$. $D_P$ contains $n_p$ positive examples $x^p$ sampled from $P(x|Y=+1)$ and $D_U$ contains $n_u$ unlabeled examples $x^u$ sampled from $P(x)$. Denote the class prior probability $\pi_p=P(Y=+1)$ and $\pi_n=P(Y=-1)$, where we follow the convention \cite{kiryo2017positive} to assume $\pi_p$ as known throughout the paper. Let $g: \mathop{\mathbb{R}}^d\rightarrow \mathop{\mathbb{R}}$ be the binary classifier and $\theta$ be its parameter, and the $L:\mathop{\mathbb{R}}\times \{+1, -1\}\rightarrow \mathop{\mathbb{R}}$ be loss function. The risk of classifier $g$, $\hat{R}_{PU}(g)$ can be approximated by

{\small
\begin{equation}
\begin{split}
    \hat{R}_{PU}(g) = \frac{\pi_p}{n_p} & \sum_{i=1}^{n_p}L(g(x^p_i), 1) + \\
    & \frac{1}{n_u}\sum_{i=1}^{n_u}L(g(x^u_i), -1) - \frac{\pi_p}{n_p}\sum_{i=1}^{n_p} L(g(x^p_i), -1), 
\end{split}
\label{eq:upu}
\end{equation}
}%
which has been known as the unbiased risk estimator for uPU \cite{du2014analysis,du2015convex,xu2017multi,elkan2008learning,xu2019positive}. It was later pointed out that the second line in Eq. (\ref{eq:upu}) would become negative due to overfitting complex models \cite{kiryo2017positive}. 

A non-negative version (nnPU) of Eq. (\ref{eq:upu}) was therefore suggested:
{\small
\begin{equation}
\begin{split}
    & \hat{R}_{PU}(g) = \frac{\pi_p}{n_p}  \sum_{i=1}^{n_p}L(g(x^p_i), 1) + \\
    & \max(0, \frac{1}{n_u}\sum_{i=1}^{n_u}L(g(x^u_i), -1) - \frac{\pi_p}{n_p}\sum_{i=1}^{n_p} L(g(x^p_i), -1))
\end{split}
\label{eq:nnpu}
\end{equation}
}%
Importance reweighting methods (e.g. uPU, nnPU) achieve the state of the arts, although treating unlabeled data as ``weighted'' negative examples still brings in unreliable supervision.
Generative adversarial networks were introduced by \cite{Hou_2018}, where the conditional generator produced both negative and positive examples resembling the unlabeled real data.
DAN \cite{liu2019discriminative} tried to recover the positive and negative distributions from the unlabeled data without requiring the class prior.
\subsection{Self-Paced Learning}
Self-paced learning \cite{NIPS2010_3923} was presented as a special case of curriculum learning \cite{bengio2009curriculum}, where the feed of training examples was dynamically generated by the model based on its learning history, aiming to simulate the learning principle of starting by learning easy instances and then gradually taking more challenging cases \cite{khan2011humans}. 
Early PU works designed heuristics for sample selection. 
In \cite{xu2019revisiting}, positive instead of negative examples are permanently selected by analyzing the distribution of sample loss. Unlike previous PU learning works which rely on crafted sample selection heuristics, we are the first to leverage the data-driven self-paced learning to progressively turn unlabeled data into labeled ones.

\subsection{Self-Supervised Learning}
In many supervision-starved fields, it is generally difficult to obtain accurate annotations, despite the vast number of unlabeled data available. Self-supervised learning aims to form ``pseudo" supervision for learning informative/discriminative features from the data, where models are required to predict on ``proxy'' tasks formed to be relevant to the target goal. It is known to benefit data-efficient learning \cite{trinh2019selfie,jing2020self}, adversarial robustness \cite{chen2020adversarial}, and outlier detection \cite{mohseni2020self}.

For example, \cite{laine2016temporal} augmented each unlabeled example with random noises, and forced consistency between the two predictions. In \cite{tarvainen2017mean}, two identical models were used during training: the student learned as usual while the teacher model generated labels and updated its weights through a moving average of the student, forcing consistency between two models. 
\cite{zhang2018deep} further suggested that, instead of exchanging examples, mutual feature distillation between peer networks can form another strong source of supervision, and can enable the collaborative learning of an ensemble of students. To our best knowledge, we are the first to consider such self-supervision to improve PU learning. 


\section{The Self-PU Framework}
\begin{figure}[ht]
 \includegraphics[scale=0.5]{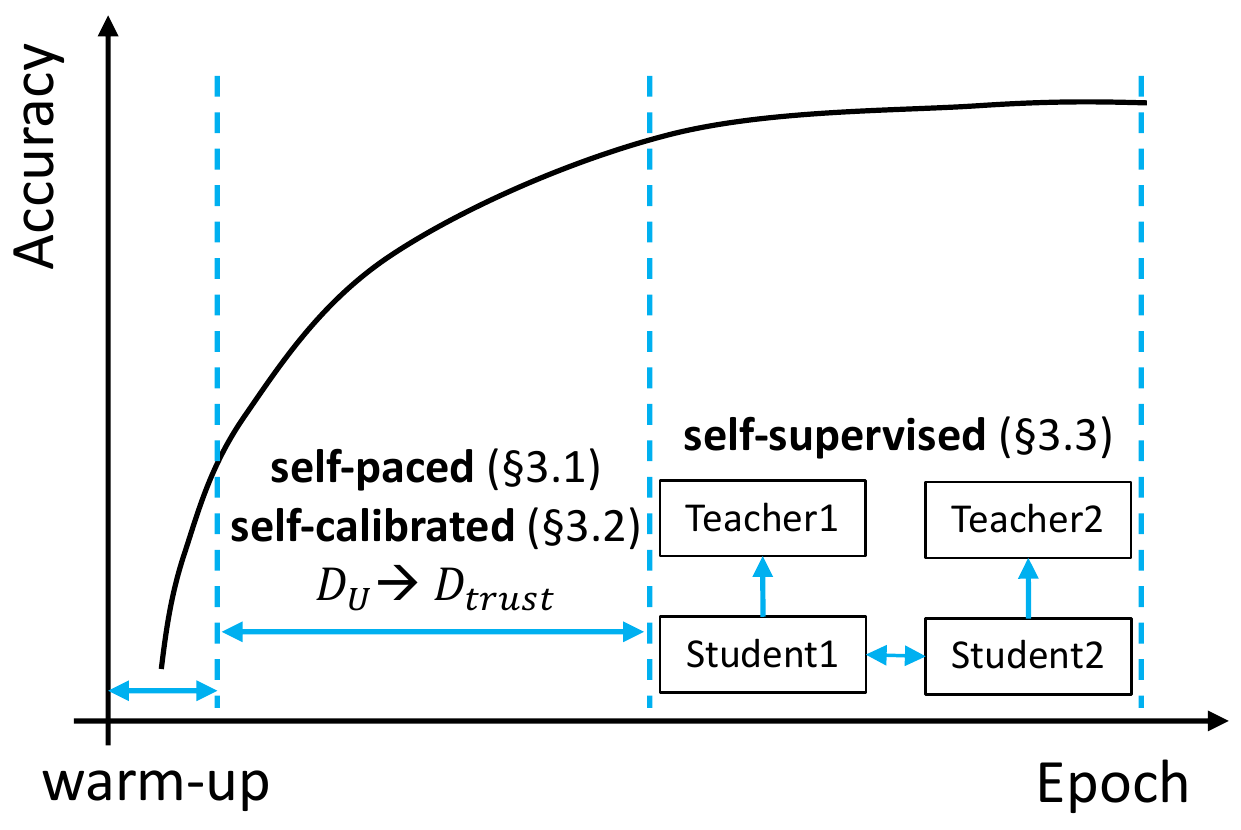}
\centering 
\caption{Illustration of the proposed ``Self-PU'' framework. After a short warm-up period, the classifier is first trained with self-paced learning, where confident examples in $D_U$ are progressively selected and labeled (positive/negative) into a trusted $D_\mathrm{trust}$ subset for supervised learning, with the loss functions over unconfident examples carefully calibrated. After collecting enough confident examples, we start the self-supervised learning via distillation between two collaborative students and their teacher networks.}
\label{fig:pipeline}
\end{figure}

Our proposed Self-PU framework exploits the learning capability of the model itself 
(Figure \ref{fig:pipeline}). We first design a self-paced learning pipeline to progressively select and label confident examples from unlabeled data for supervised learning. On top of that, we calibrate the loss functions over the unconfident examples via meta-learning. Moreover, a consistency loss is introduced between peer networks with different learning paces, which collaboratively teach each other. We further extend our consistency from peer networks to their moving-averages \cite{tarvainen2017mean, laine2016temporal}, as another form of supervision.

\subsection{Self-Paced PU Learning}
Despite the success of unbiased PU risk estimators, they still rely on the estimated class prior and reduced weights on unlabeled data. As shown in \cite{arpit2017closer}, during gradient descent, deep neural networks tend not to memorize all training data at the same time but tend to memorize frequent or easy patterns first and later irregular patterns. If we could first find out easy examples and label them with confidence, and then augmenting this labeled pool for the training progress, then we can enjoy ``progressively increased'' confident full supervision along with the training, in addition to the weak supervision from the PU risk estimators.


Given the model $g$, an input example $x$ and the corresponding label $y$, we may compute the output $g(x)$ and then calculate the probability of $x$ being positive as $p(x) = P(Y=+1|x) = f(g(x))$, where $f$ is a monotonic function of mapping $\mathop{\mathbb{R}} \rightarrow [0,1]$ (\textit{e.g.} sigmoid function). A greater $p(x)$ suggests higher confidence that $x$ belongs to positive class as predicted by $g$, and vice versa. By descending sort of $p(x)$ each time, we can select $n$ most confident positive and $n$ most confident negative examples from the current unlabeled data pool $D_U$. They will be removed from $D_U$ and added to our trusted subset $D_\mathrm{trust}$, considered as labeled training examples hereinafter.

Let $L_\mathrm{CE}(x, y)$ be the cross entropy loss: 
{\small
$$L_\mathrm{CE}(x, y) = \log f(g(x)) \mathbb{I}_{y=1} + \log (1 - f(g(x))\mathbb{I}_{y=-1}$$
}, $L_\mathrm{nnPU}(x, y)$ be the nnPU risk with Sigmoid loss, and together with the given positive subset $D_P$, our hybrid loss for self-paced learning becomes:
{\small
\begin{equation}
\begin{split}
    L_\mathrm{SP} = \sum_{(x, y)\in  D_\mathrm{trust}}L_\mathrm{CE}(x, y) + 
    \sum_{x\in D - D_\mathrm{trust}}L_\mathrm{nnPU}(x)
\end{split}
\label{eq:loss_sp}
\end{equation}
}%
Note that previous works select either only confident positive examples \cite{xu2019revisiting} or negative examples \cite{li2003learning}, while our self-paced learning selects both. Since the cross entropy is used as our supervised loss, one advantage is that the trusted sets of positive/negative samples are balanced in size at each sampling step,
avoiding the potential pitfall of extreme class imbalance caused by incrementally sampling only one class.

Besides, previous sample selection \cite{xu2019revisiting} often sticks to a pre-fixed learning schedule. In contrast, we unleash more flexibility for the model to automatically and adaptively adjust its own learning pace, via the following techniques. Later on, we will experimentally verify their effectiveness via a step-by-step ablation study. 

\subsubsection{Dynamic Rate Sampling}
As the learning progresses, training examples with easy/frequent patterns and those with harder/irregular patterns are memorized in different training stages \cite{arpit2017closer}. It is important to make our self-paced learning compatible with the memorizing process of the model. A small number of easy examples should be selected first, and then intermediate to hard examples can be labeled after the model is well-trained. Instead of fixing the number of selected confident examples, we propose to dynamically choose the number of confident examples during the self-paced learning.
Specifically, as the self-paced learning proceeds, we linearly increase the size of $D_\mathrm{trust}$ from 0 to $r|D_U|$, where the sampling ratio $r$ could range from 10\% to 40\% in our experiments (see section~\ref{sec:student_pace}).
Empirically, we first ``warm-up'' the model by training 10 epochs before starting the self-paced learning, in order to keep the selected confident examples as accurate as possible.


\subsubsection{``In-and-Out'' Trusted Set}
In previous sample selection approaches, once selected, a trusted example will never be deprived of its label during the subsequent training. In contrast, we allow our training to ``regret'' on earlier selections in $D_\mathrm{trust}$. Especially at the early training stage, the intermediate model might not be well-trained enough and not always reliable for predicting labels, which could mislead training if continuing acting as supervision. To this end, we adaptively update $D_\mathrm{trust}$ by also re-examining its current examples each time when we augment new confident ones. The previously selected examples will be removed from $D_\mathrm{trust}$ if their predictions by the current model become of low-confidence, and will be treated as unlabeled again.


\subsubsection{Soft Labels}\label{sec:soft_label}
Instead of giving the selected confident examples hard labels, we directly use the prediction $f(g(x))$ as soft labels: $[1 - f(g(x)), f(g(x))]$ as $[P(Y=-1|x), P(Y=+1|x)]$, as the practice of label smoothing \cite{Szegedy_2016} appears to benefit learning robustness against label noise. 

\subsection{Self-Calibrated Loss Reweighting}\label{sec:reweighting}
Only leveraging nnPU risk on $D_U - D_\mathrm{trust}$ may not be optimal, as some examples in this set can still provide meaningful supervision. To exploit more supervision from this noisy sets, we introduce a \textit{learning-to-reweight} paradigm \cite{ren2018learning} to the PU learning field for the first time. Letting $L_\mathrm{CES}(x) = f(g(x))\log f(g(x)) + (1 - f(g(x))) \log (1 - f(g(x)))$ be the cross entropy loss function with soft labels in Sec. \ref{sec:soft_label}, we adaptively combine $L_{\mathrm{CES}}$ and $L_{\mathrm{nnPU}}$ for each example $
x_i$ in a batch from $D_U - D_\mathrm{trust}$, namely:
$$
l(x_i, \bm{w}_i) = w_{i,1} L_\mathrm{CES}(x_i) + w_{i,2} L_\mathrm{nnPU}(x_i)
$$

Let $n$ be the mini-batch size. To learn the optimal $\bm{w} = [\bm{w}_1, \bm{w}_2, \dots, \bm{w}_n]^T$ together in training, we update the model $g$ for a single gradient descent step on $l$ with $\bm{w_i}$ very small (\textit{i.e.} a perturbation) of a mini-batch of training examples w.r.t. parameters of models $\theta$, followed by a gradient descent step on the cross entropy loss of a mini-batch of validation examples w.r.t $\bm{w}$, and then rectify the output to be non-negative. The procedure is described as follows: 
\begin{align}
    \theta^* &= \theta - \delta \nabla \sum_{i=1}^{n} l(x_i,y_i,\bm{w}_i) \\
    \bm{u}_i &= -\frac{\partial}{\partial \bm{w}_i} \frac{1}{m} \sum_{j = 1}^m L^*_{\mathrm{CE}}(x_j^v, y_j^v)|_{\bm{w}_i=\bm{0}}\\
    \tilde{\bm{w}_i} &= \max (\bm{u}_i, \bm{0}), 
    w_{i, 1} = \frac{\tilde{w}_{i, 1}}{\sum_{i} \tilde{w}_{i, 1}}, w_{i, 2} = \frac{\tilde{w}_{i, 2}}{\sum_{i} \tilde{w}_{i, 2}}
\end{align}
where $\delta$ denotes the step size, $m$ the mini-batch size on the validation set which contains clean positive and negative examples, and $(x_j^v, y_j^v), j = 1, 2, \dots, m$ an example from the validation set with the ground-truth label. $L^*_\mathrm{CE}(x, y)$ calculates loss using the updated parameters $\theta^*$. 

Meanwhile, on $D_U - D_{trust}$, weighting the cross-entropy loss too much might not be beneficial to the classifier, especially when the soft labels are not accurate enough. 

Therefore, we restrict the total weights of the cross-entropy loss via a balancing factor $\gamma$:
{\small
\begin{align}
    &T = \sup\{k:\sum_{i = 1}^k w_{2,i} < \gamma n \} \\
    &w^*_{i,1} = w_{i,1} \mathrm{1}_{ \{i<T\} }, w^*_{i,2} = w_{i,2} \mathrm{1}_{ \{i<T\} } + \mathrm{1}_{ \{i\ge T\} }
\end{align}
}%

The corresponding hybrid loss becomes:
{\small
\begin{align}
    &L_\mathrm{SP+Reweight} = \sum_{(x, y)\in  D_\mathrm{trust}}L_\mathrm{CE}(x, y) + \\
    &\sum_{\bm{x}\subset (D_U - D_\mathrm{trust})}l^*(\bm{x}) + \sum_{x\in D_P}L_\mathrm{nnPU}(x),
    \label{eq:loss_sp_reweight}
\end{align}
}%
where $l^*(\bm{x}) = \frac{\sum_{i=1}^n l(x_i, \bm{w}^*_i)}{n}$.
\subsection{Self-Supervised Consistency via Distillation}
To explore additional sources of supervision, we encourage \textit{two forms of self-supervised consistency}: among different learning paces of the model, and along the model's own moving averaged trajectory. The two goals are altogether achieved by an innovative distillation scheme, with a pair of collaborative student models and their teacher models.

\subsubsection{Consistency for Different Learning Paces: Making A Pair of Students}
The consistency between two self-paced models trained with different paces (\textit{i.e.} sampling ratio in self-paced learning) makes the trained model more resilient to perturbations caused by training stochasticity. To form this self-supervision, we simultaneously train two networks that share the identical architecture, with the same $D_P$ and $D_U$ to start on. However, they are set with different confidence thresholds and select different amounts from $D_\mathrm{U}$ to $D_\mathrm{trust}$ each time, making their learning paces un-synchronized and resulting in two different ``trusted sets'' $D_\mathrm{trust1}$ and $D_\mathrm{trust2}$. Since two students' estimations of the probabilities of classes may differ, we force the consistency between two students as a source of distillation, via a mean-square-error (MSE) loss on two models' predictions. 

Denote two networks $g_1$ and $g_2$,
the MSE from $g_1$ to $g_2$ is defined over $D-D_\mathrm{trust1}$:
{\small 
\begin{equation}
    L_\text{MSE}(g_1, g_2 ,x) = ||f(g_1(x))-f(g_2(x))||^2, x\in D-D_\mathrm{trust1}
\label{eq:mse1}
\end{equation}
}%
The MSE from $g_2$ to $g_1$ is defined over $D-D_\mathrm{trust2}$:
{\small
\begin{equation}
    L_\text{MSE}(g_2, g_1, x) = ||f(g_2(x))-f(g_1(x))||^2, x\in D-D_\mathrm{trust2}
\label{eq: mse2}
\end{equation}
}%


A \textit{hard sample mining} strategy is further adopted on top, where we only calculate MSE over those ``challenging'' unlabeled examples whose nnPU risks (\ref{eq:nnpu}) are large. 

Later on, the pair of networks will become two student models for distillation. We therefore have our \textit{first part} of self-supervised consistency loss as:

{\small
\begin{equation}
\begin{split}
L_\mathrm{students} = 
&\sum_{x\in D-D_\mathrm{trust1}}l_\text{stu}(g_1, g_2, x) +\\& \sum_{x\in D-D_\mathrm{trust2}}l_\text{stu}(g_2, g_1, x),
\end{split}
\end{equation}
where 
\begin{equation}
\begin{split}
&l_\text{stu}(g_1, g_2, x) = \\
&\begin{cases}
    L_\text{MSE}(g_1, g_2, x),& L_\mathrm{nnPU}(x) > \alpha L_\text{MSE}(g_1, g_2, x) \\
    0, & L_\mathrm{nnPU}(x) \leq \alpha L_\text{MSE}(g_1, g_2, x)
\end{cases}
\end{split}
\label{eq:mutual_learning}
\end{equation}}%

We study the effect of choosing $\alpha$ in section \ref{sec:alpha}. The mean squared error between the two students is only calculated on $D-D_\mathrm{trust1}$ and $D-D_\mathrm{trust2}$. One reason why we choose such design is that the accuracy on $D_\mathrm{trust1}$ and $D_\mathrm{trust2}$ discovered by the self-paced learning is much higher than the accuracy on $D$. In addition, on the $D_\mathrm{trust1}$ and $D_\mathrm{trust2}$ the prediction entropy is 0.005, while on the unlabeled set it is 0.074, which indicates much lower confidence. 

\subsubsection{Consistency for Moving Averaged Weights: Adding Teachers to Distill}
Inspired by \cite{tarvainen2017mean}, in addition to the consistency between two students, we also encourage them to be consistent with their moving averaged trajectory of weights.
Assume that $g_1$ and $g_2$ are parameterized by $\theta_1$ and $\theta_2$. For each student we introduce a new teacher model, $G_1$ and $G_2$, parameterized by $\Theta_1$ and $\Theta_2$ with the same structure as $g_1$ and $g_2$. The weights of $G_1$, $G_2$ are updated via the following moving average:
\begin{equation}
\begin{split}
    \Theta_{1,t} = \beta\theta_{1,t-1} + (1-\beta)\theta_{1,t} \\
    \Theta_{2,t} = \beta\theta_{2,t-1} + (1-\beta)\theta_{2,t}
\end{split}
\label{eq:teacher}
\end{equation}
where $\theta_{1,t}$ denotes the instance of $\theta_{1}$ at time $t$, and similarly for others.
We study the effect of $\beta$ in section \ref{sec:beta}.

An MSE loss is next enforced for $G_1$ and $G_2$ to distill from $g_1$ and $g_2$, namely:
{\small
\begin{equation}
\begin{split}
    L_\mathrm{teachers} =& \sum_{x\in D}||f(G_1(x)) - f(
    g_1(x))||^2 \\
    &+ \sum_{x\in D}||f(G_2(x)) - f(g_2(x))||^2
\end{split}
\label{eq:mean_teacher}
\end{equation}
}%
The above constitutes the \textit{second part} of our self-supervised consistency cost. 

In summary, the benefits of self-supervised learning for PU learning come from two folds: 1) the enlarged labeled examples $(D_\text{trust})$ introduces stronger supervision into PU learning and brings high accuracy; 2) the consistency cost between diverse student and teacher models introduces the learning stability (low variance). Eventually, our overall loss function\footnote{Since here we have two students of different learning paces, our $L_\mathrm{SP + Reweight}$ is also extended to both $D_\mathrm{trust1}$ and $D_\mathrm{trust2}$.} of Self-PU is:
{\small
\begin{equation}
\begin{split} 
    L = L_\mathrm{SP + Reweight} + L_\mathrm{students} + L_\mathrm{teachers}.
\label{eq:loss_all}
\end{split}
\end{equation}
}%
In all experiments and as shown in Figure \ref{fig:pipeline}, we first apply self-paced learning and self-calibrated loss reweighting from the 10th epoch to the 50th epoch, followed by a self-distillation period from 50th to 200th epoch. That allows for the models to learn sufficient meaningful information before being distilled. After training, we compare the validation accuracy of two teacher models and select the better performer to be applied on the testing set\footnote{Note that, here we only select one and discard the other, only for simplicity purpose. Other approaches, such as average or weighted-fusion of the two teachers models, are applicable too.}.

\section{Experiments}
\subsection{Datasets}
In order to evaluate our proposed ``Self-PU'' learning framework, we conducted experiments on two common testbeds for PU learning: MNIST, and CIFAR-10; plus \textbf{a new real-world benchmark}, \textit{i.e.} ADNI \cite{jack2008alzheimer}, for the application of Alzheimer's Disease diagnosis.
\subsubsection{Introduction to the ADNI database}

\begin{figure}[h!]
\includegraphics[scale=0.3]{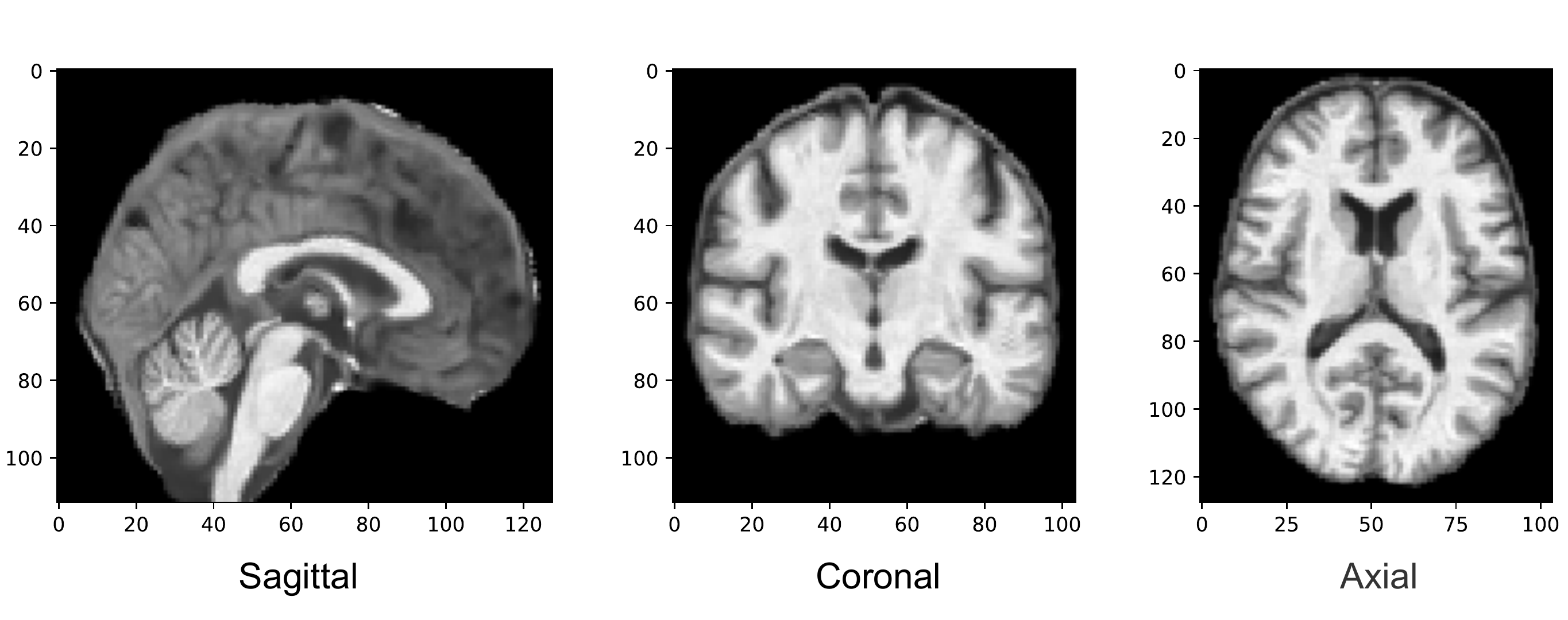}
\centering
\caption{Cross-sectional imaging of a 104$\times$128$\times$112 MRI example from the ADNI dataset. Images are from the $52^{nd}$, $64^{th}$, $56^{th}$ slice of sagittal, coronal, and axial plane, respectively. An MRI image is of gray scale with a value from 0 to 255 for each voxel, and was processed by the intensity inhomogeneity correction, skull-stripping and cerebellum removing.}
\label{adni_visual}
\end{figure}

The Alzheimer’s Disease Neuroimaging Initiative (ADNI) database\footnote{\url{http://adni.loni.usc.edu}} was constructed to test whether brain scans, \textit{e.g.} magnetic resonance imaging (MRI) and other biological markers, can be utilized to predict the early-stage Alzheimer’s disease (AD), in order for more timely prevention and treatment. The dataset, especially its MRI image collection, has been widely adopted and studied for the classification of Alzheimer’s disease \cite{khvostikov20183d,li2015robust}. Fig. \ref{adni_visual} shows visual examples.

\begin{table*}[t!]
\centering
\caption{Specification of benchmark datasets and models.}
\label{dataset}
\begin{tabular}{ccccccc}\toprule
Dataset & \#Train & \#Test & Input Size & $\pi_p$ & Positive/Negative & Model \\ \midrule
MNIST & 60,000 & 10,000 & 28$\times$28 & 0.49 & Odd/Even & 6-layer MLP \\
CIFAR-10 & 50,000 & 10,000 & 3$\times$32$\times$32 & 0.40 & Traffic tools/Animals & 13-layer CNN \\
ADNI & 822 & 113 & 104$\times$128$\times$112 & 0.43 & AD Positive/Negative & 3-branch 2-layer CNN \\ \bottomrule
\end{tabular}
\end{table*}

Traditionally, the machine learning community considers the AD diagnosis task as a binary, fully supervised classification task, between the \textit{patient} and the \textit{healthy} classes. It has \textbf{never} been connected to PU learning. Yet, we advocate that this task could become a new \textbf{suitable, realistic} and \textbf{challenging} application benchmark for PU learning. 

The early-stage AD prediction/diagnosis is highly nontrivial for multi-fold, field-specific reasons. \underline{First}, many nuance factors can heavily affect the feature effectiveness, ranging from individual patient variability to (mechanical/optical) equipment functional fluctuations, to manual operation and sensor/environment noise. \underline{Second}, within the whole-brain scans, only some (not fully-specified) local brain regions are found to be indicative of AD symptoms.  
\underline{Third and most importantly}, in contrast to the diagnosed patients, the remaining population, who are not yet clinically diagnosed with AD, \textit{cannot} be simply treated as all \textit{healthy}: on one hand, the above challenges of AD diagnosis inevitably lead to incorrectly missed patient cases; on the other hand, and \textbf{more notably}, the AD patients go through a stage called \textit{mild cognitive impairment} (MCI) \cite{larson2004survival,duyckaerts1997diagnosis}, a critical transition period between the expected cognitive decline of normal aging, and the severe decline of true dementia. During the MCI stage, those people were ``clinically'' considered as AD patients already (if diagnosed with more intrusive bio-chemical means); however, no symptom is known to be observable in current MRI images or other bio-markers. 

In other words, the MCI examples have definitely been included in the currently ``healthy''-labeled samples in ADNI, while they should have belonged to the ``patient'' class. In training, we label patients as \textbf{positive} class, the ``healthy''-labeled examples can then be  considered as \textbf{unlabeled} class, which mixes the true healthy people (\textit{i.e.}, from the actual \textbf{negative} class) and the MCI people (\textit{i.e.}, from the \textbf{positive} class). We communicated with several seasoned medical doctors practicing in AD fields, and \textbf{they unanimously agreed} that AD diagnosis should be described as a PU learning problem rather than (the traditional treatment as) a binary classification problem. In this paper, we study the specific setting of MRI image classification task on the ADNI dataset, while other bio-marker classification can be similarly studied in PU settings. 

\subsubsection{Dataset Setting}
We report our dataset protocols towards PU learning. More metadata are summarized in Table \ref{dataset}.
\begin{itemize}
    \item \textbf{MNIST}: odd numbers 1, 3, 5, 7, 9 form the positive class while even numbers 0, 2, 4, 6, 8 form the negative.
    \item \textbf{CIFAR-10}: four vehicles classes (``airplane'', ``automobile'', ``ship'', ``truck'') constitute the positive class, and six animal classes (``bird'', ``cat'', ``deer'', ``dog'', ``frog'' ``horse'') constitute the negative.
    \item \textbf{ADNI}:
    We utilized the public ADNI data set as \cite{li2015robust,yuan2018feasibility} suggested:
    The T1-weighted MRI images were processed by first correcting the intensity inhomogeneity, followed by skull-stripping and cerebellum removing. We consider the subjects as positive class if they: 1) have positive clinical diagnosis records on file; or 2) have their standardized uptake value ratio (SUVR) values \footnote{SUVR is a therapy monitoring or response, considered as an important indicator of Alzheimer's Disease.} no less than 1.08 \cite{villeneuve2015existing,ott2017brain,yuan2018feasibility}. While this estimate can be treated as ``golden rule'' in clinical practice and is shown to work well in our experiments (Table \ref{table:sota_adni}), it can be further adjusted flexibly and used in our framework with ease.
\end{itemize}
Following the convention of nnPU \cite{kiryo2017positive}, we use $n_p = |D_P| = 1000$ in MNIST and CIFAR-10.
 In ADNI, we end up with $n_p = 113$. $n_u = |D_U|$ equals the size of remaining training data on all three datasets. $\pi_p$ is the proportion of true positive examples in the dataset. 

\subsection{Baselines and Implementations}

Following nnPU \cite{kiryo2017positive}, we used a 6-layer \textit{multilayer perceptron} (MLP) with ReLU on MNIST. On CIFAR-10, we use a 13-layer CNN with ReLU. We design a multi-scale network for ADNI, which is used as the backbone for all compared baselines: please see supplementary materials for details. 
We use Adam optimizer with a cosine annealing learning rate scheduler for training. The batch size is 256 for MNIST and CIFAR-10, and 64 for ADNI. The $\gamma$ is set to $\frac{1}{16}$. The batch size of validation examples equals to the batch size of the training examples. 

For a fair comparison, each experiment runs five times, 
and the mean and standard deviations of accuracy are reported.


\subsection{Ablation Study}
In this section, we carry out a thorough ablation study, on the key components introduced during the self-paced learning stage (\textit{i.e.}, the selection of trusted set) and the self-supervised distillation stage (\textit{i.e.}, the diversity of students, the effect of hard sample mining when training students, and the effect of weight averaging further by teachers). All experiments are conducted on the \textbf{CIFAR-10} dataset\footnote{To conduct controlled experiments we disable the self-calibration strategy in Table \ref{table:sp_pace}, \ref{table:alpha}, \ref{table:beta}}. We will study the effect of $\gamma$ in the supplementary materials. 

\subsubsection{Selection of ``Trusted Set'' $D_\text{trust}$}
Since self-paced learning aims to mine more confident positive/negative examples, it is crucial to ensure the trustworthiness of the selected $D_\text{trust}$.  We therefore calculate the accuracy of assigned labels for $D_\text{trust}$ along the self-paced training, as an indicator of the sampling strategy reliability. 

We compare three settings: 1) \textit{``Fixed sampling size''}: each time, the model selects a fixed number of samples (\textit{e.g.} 25\% of examples in $D_U$), assigning soft labels and adding them into $D_\text{trust}$. Meanwhile, low-confidence samples in $D_\text{trust}$ will also be removed in next round of selection. 2) \textit{``Sampling without replacement''}: each example selected by model will permanently reside in $D_\text{trust}$. Here the size of $D_\text{trust}$ is linearly increased along the training progress. 3) Our \textit{default approach} in Self-PU: both ``Dynamic Rate Sampling'' and ``In-and-Out Trusted Set'' are enabled. All three settings end up with $|D_\text{trust}| = 0.25|D_U|$.

\begin{figure}[ht]
\includegraphics[scale=0.4]{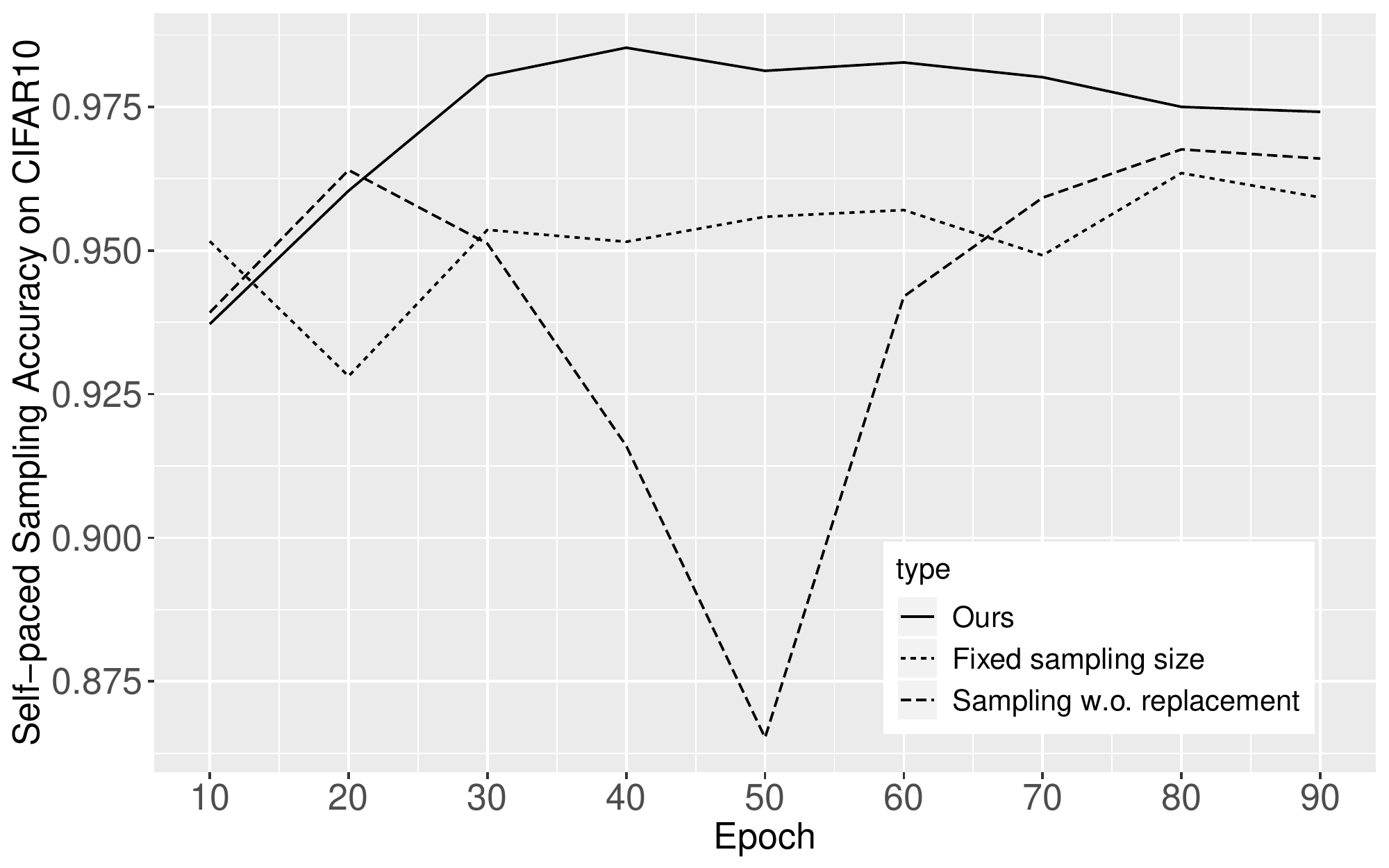}
\centering 
\caption{Accuracy of selected confident examples during self-paced learning. We compared self-paced learning with three different sampling settings: fixed sampling size (dot line), sampling without replacement (dash line), and our proposed dynamic ``in-and-out'' sampling (solid line). It is clear that self-paced learning with fixed sampling size or without replacement suffers from low sampling accuracy, and no-replacement is even jeopardized by the inaccurate examples remain in the $D_\text{trust}$.}
\label{self_paced_sampling_acc}
\end{figure}

From Figure \ref{self_paced_sampling_acc}, we clearly see that sampling either with a fixed size or without replacement results in a less reliable selection of $D_\text{trust}$, compared to our strategy. Moreover, the inaccurately selected examples in $D_\text{trust}$ will further cause much unstable training (dash line). 
We demonstrate that both ``Dynamic Rate Sampling'' and ``In-and-Out Trusted Set'' are vital to achieving an accurate and stable self-paced learning (solid line). Table \ref{table:ablation0} shows the final test accuracy of three settings, where our proposed self-paced learning pipeline ($L_\text{SP}$) significantly outperforms the other two settings ($L_\text{SP}$ of fixed sampling size and sampling without replacement). The better accuracy and lower variance show the advantage of our strategy.

\begin{table}[t!]
\centering
\caption{Classification comparison on CIFAR-10: we report both \textbf{means} and \textbf{standard deviations} (in parentheses) from five runs. $L_\text{SP}$: self-paced training. $L_\text{SPS}$: self-paced training with soft label in Sec. \ref{sec:soft_label}. $L_\text{SP + Reweighting}$: self-paced training with self-calibrated loss reweighting in section \ref{sec:reweighting}. $L_\text{SPS + Reweighting}$: self-paced training with soft label and self-calibrated loss reweighting in section \ref{sec:reweighting}. $L_\text{students}$: self-distillation from a pair of students. $L_\text{teacher}$: self-distillation from teacher networks. Self-PU: $L_\text{SPS + Reweighting} + L_\text{students}+L_\text{teacher}$ }
\begin{tabular}{cc}
\toprule
Method & CIFAR-10 \% \\ \midrule
nnPU (baseline) & 88.60 (0.40) \\ \midrule
$L_\text{SP}$ (fixed size)  & 88.05 (0.59) \\
$L_\text{SP}$ (w.o. replacement)  & 88.27 (0.43) \\
$L_\text{SP}$  & 88.66 (0.40) \\
$L_\text{SPS}$ &  88.75 (0.27)\\
$L_\text{SP + Reweighting}$  &  89.25 (0.42) \\
$L_\text{SPS + Reweighting}$ &  89.39 (0.36) \\
\midrule
$L_\text{SP} + L_\text{students}$ & 88.84 (0.36) \\
$L_\text{SPS} + L_\text{students}$ & 88.93 (0.28) \\
$L_\text{SP} + L_\text{students} + L_\text{teacher}$ & 89.43 (0.42) \\
$L_\text{SPS} + L_\text{students} + L_\text{teacher}$ & 89.65 (0.33) \\
\midrule
Self-PU & \textbf{89.68 (0.22)} \\
\bottomrule
\end{tabular}\label{table:ablation0}
\end{table}

\begin{table}[!htb]
\centering
\caption{Study of student diversity (learning paces) for two-student distillation on CIFAR-10 dataset. Pace1/Pace2 denotes the final ratio of $|D_\text{trust}|$ over $|D_U|$.}
\vspace{0.5em}
\begin{tabular}{ccc}
\toprule
Pace1 & Pace2 & Test Accuracy \% \\ \midrule
10\% & 40\% & 89.32 (0.36)\\
15\% & 35\% & 89.55 (0.46)\\
\textbf{20\%} & \textbf{30\%} & \textbf{89.65 (0.33)}\\
25\% & 25\% & 89.64 (0.47)\\
\bottomrule
\end{tabular}\label{table:sp_pace}
\label{table:student_pace}
\end{table}

\begin{table}[!htb]
\centering
\caption{Study of hard sample mining threshold $\alpha$ for two-student distillation on CIFAR-10 dataset. Smaller $\alpha$ indicates stronger distillation (Eq.~(\ref{eq:mutual_learning}))}
\vspace{0.5em}
\begin{tabular}{cc}
\toprule
$\alpha$ & Test Accuracy \% \\ \midrule
5 & 89.59 (0.39)\\
\textbf{10} & \textbf{89.65 (0.33)}\\
20 & 89.38 (0.51)\\
\bottomrule
\end{tabular}\label{table:alpha}
\end{table}

\begin{table}[!htb]
\centering
\caption{Study of smoothing coefficient $\beta$ for teacher networks on CIFAR-10 dataset. Greater $\beta$ indicates slower updates of teachers from the students (Eq.~(\ref{eq:mean_teacher}))}
\vspace{0.5em}
\begin{tabular}{cc}
\toprule
$\beta$ & Test Accuracy \% \\ \midrule
0.2 & 89.37 (0.39)\\
\textbf{0.3} & \textbf{89.65 (0.33)}\\
0.4 & 89.47 (0.41)\\
\bottomrule
\end{tabular}\label{table:beta}
\end{table}

\begin{table}[!htb]
\centering
    \caption{Study of $\gamma$ for self-calibrated loss reweighting on CIFAR-10. Greater $\gamma$ indicates larger weight on cross-entropy (Eq.~(\ref{eq:loss_sp_reweight}))}
    \vspace{0.5em}
    \begin{tabular}{cc}
        \toprule
        $\gamma$ & Accuracy \\
        \midrule
        0.125 & 89.29\% \\
        0.100 & 89.42\% \\
        0.075 & 89.55\% \\
        \textbf{0.063} & \textbf{89.68}\% \\
        0.050 & 89.67\% \\
        0.000 & 89.65\% \\
        \bottomrule
    \end{tabular}
    \label{tab:gamma}
\end{table}

\begin{table*}[!htb]
\footnotesize
\begin{minipage}[t]{.63\linewidth}
\centering
\caption{Classification comparison on MNIST and CIFAR-10.
``\textbf{*}'' indicates that 3,000 positive examples were initialized for training, while others used 1,000.}
\begin{tabular}{ccc}
\toprule
Method & MNIST \% & CIFAR-10 \%\\ \midrule
uPU \cite{du2014analysis} & 92.52 (0.39) & 88.00 (0.62) \\
nnPU \cite{kiryo2017positive} & 93.41 (0.20) & 88.60 (0.40) \\
DAN* \cite{liu2019discriminative} & - & 89.7 (0.40) \\ \midrule
Self-PU & 94.21 (0.54) &  89.68 (0.22) \\
$\text{Self-PU}^*$ & \textbf{96.00 (0.29)} & \textbf{90.77 (0.21)} \\
\bottomrule
\end{tabular}\label{table:sota}
\end{minipage}%
\hfill%
\begin{minipage}[t]{.33\linewidth}
\centering
\caption{Classification accuracy of different methods on ADNI. ``naive'' means that we treat the entire unlabeled class as negative.}
\begin{tabular}{cc}
\toprule
Method  & ADNI \%\\ \midrule
naive & 73.27 (1.45) \\
uPU & 73.45 (1.77) \\
nnPU & 75.96 (1.42) \\ \midrule
Self-PU & \textbf{79.50 (1.80)} \\
\bottomrule
\end{tabular}\label{table:sota_adni}
\end{minipage}%
\end{table*}

\subsubsection{Effects of Student Diversity}
\label{sec:student_pace}
Different learning paces enable the diversity of two students and thus make the collaborative teaching between two students effective. Therefore we study how the student diversity, \textit{i.e.} combination of their different learning paces, can affect the final results. Table \ref{table:sp_pace} considers three different pace pairs. For example, Pace1 ``10\%'' means that the self-paced learning of the first student model will end up with $|D_\text{trust}| = 0.1|D_U|$, and all students will complete the sampling for self-paced learning within the same number of training epochs. 
Table \ref{table:sp_pace} shows that, while student diversity helps (``20\%'' + ``30\%'' $>$ ``25\%'' + ``25\%''), too large student pace discrepancy will hurt the learning too (``20\%'' + ``30\%'' $>$ ``10\%'' + ``40\%''). Students with very different paces are harmful because a large gap in two learning paces results in a smaller intersection-over-union of $D_\mathrm{trust1}$ and $D_\mathrm{trust2}$. It is difficult to keep consistency between two models trained with different amounts of labeled data. Therefore, it is important to keep diversity, while not too extreme.

\subsubsection{Effects of Sample Mining Threshold}\label{sec:alpha}

$L_\text{students}$ takes the hard sample mining threshold $\alpha$ as an important hyperparameter: the smaller $\alpha$ is, the more examples are counted in computing the mean squared error, which implies stronger self-supervision consistency between two students. Table \ref{table:alpha} shows that a moderate $\alpha$ = 10 leads to the best performance. Understandably, either ``under-mining'' ($\alpha$ = 20) or ``over-mining'' ($\alpha$ = 5) hurts the performance: the former is not sufficiently regularized, while the latter starts to eliminate the emphasis over hard examples.



\subsubsection{Effects of Smoothing Coefficient $\beta$}\label{sec:beta}
The smoothing coefficient $\beta$ controls how ``conservative'' we distill the teachers from the students: the larger $\beta$ is, the more reluctant the teacher models get updated from the students. Table \ref{table:beta} investigate three different $\beta$ values: similar to the previous experiment on $\alpha$, $\beta$ also favors a reasonably moderate value, while either overly large or small $\beta$, corresponding to ``over-smoothening'' and ``under-smoothening'' during the distillation from students to teachers respectively, degrades the final performance. 

\subsubsection{Effects of $\gamma$ for Self-Calibrated Loss Reweighting}
The balancing factor $\gamma$ restricts the total weights of the cross-entropy loss. In Table \ref{tab:gamma}, we report the validation accuracy with different $\gamma$, and we can see that the optimal choice of $\gamma$ is 0.063. Table \ref{tab:gamma} indicates that mining examples with our calibrated loss contribute to better supervision than using only $L_\mathrm{nnPU}$, while too much weight on $L_\mathrm{XE}$ may lead to worse validation accuracy.


\subsubsection{Effect of Teacher and Students}
We verify the effect of two types of distillation in our self-supervised learning: mutually between $L_\text{students}$, and by $L_\text{teachers}$. In Table \ref{table:ablation0}, distillation from two students with different learning paces ($L_\text{students}$) improve the accuracy of nnPU baseline from 88.60\% to 88.84\% on CIFAR-10. By adding two teachers for self-distillation, the performance is further boosted to 89.43\%, which endorses the complementary power of two types of self-distillation.

\subsection{Comparison to State-of-the-Art Methods}

\subsubsection{Results on the MNIST and CIFAR-10 Benchmarks}

We compare the performance of the proposed Self-PU with several popular baselines: the unbiased PU learning (uPU) \cite{du2014analysis}; the non-negative PU learning (nnPU) \cite{kiryo2017positive}\footnote{We reproduced the uPU and nnPU baselines using the official codebase from: \url{https://github.com/kiryor/nnPUlearning}}, and DAN (a latest GAN-based PU method) \cite{liu2019discriminative}. 

Table \ref{table:sota} summarizes the comparison results on MNIST and CIFAR-10. On MNIST, Self-PU outperforms uPU and nnPU by over 0.5\%, setting new performance records. On CIFAR-10, Self-PU surpasses nnPU by over 1\% (a considerable gap). More importantly, by only leveraging 1000 positive examples, Self-PU achieves comparable performance as DAN where 3000 positive samples were used. Training with 3000 positive examples further boosts our performance which outperforms DAN by 1\%.  


\begin{figure}[ht]
\includegraphics[scale=0.35]{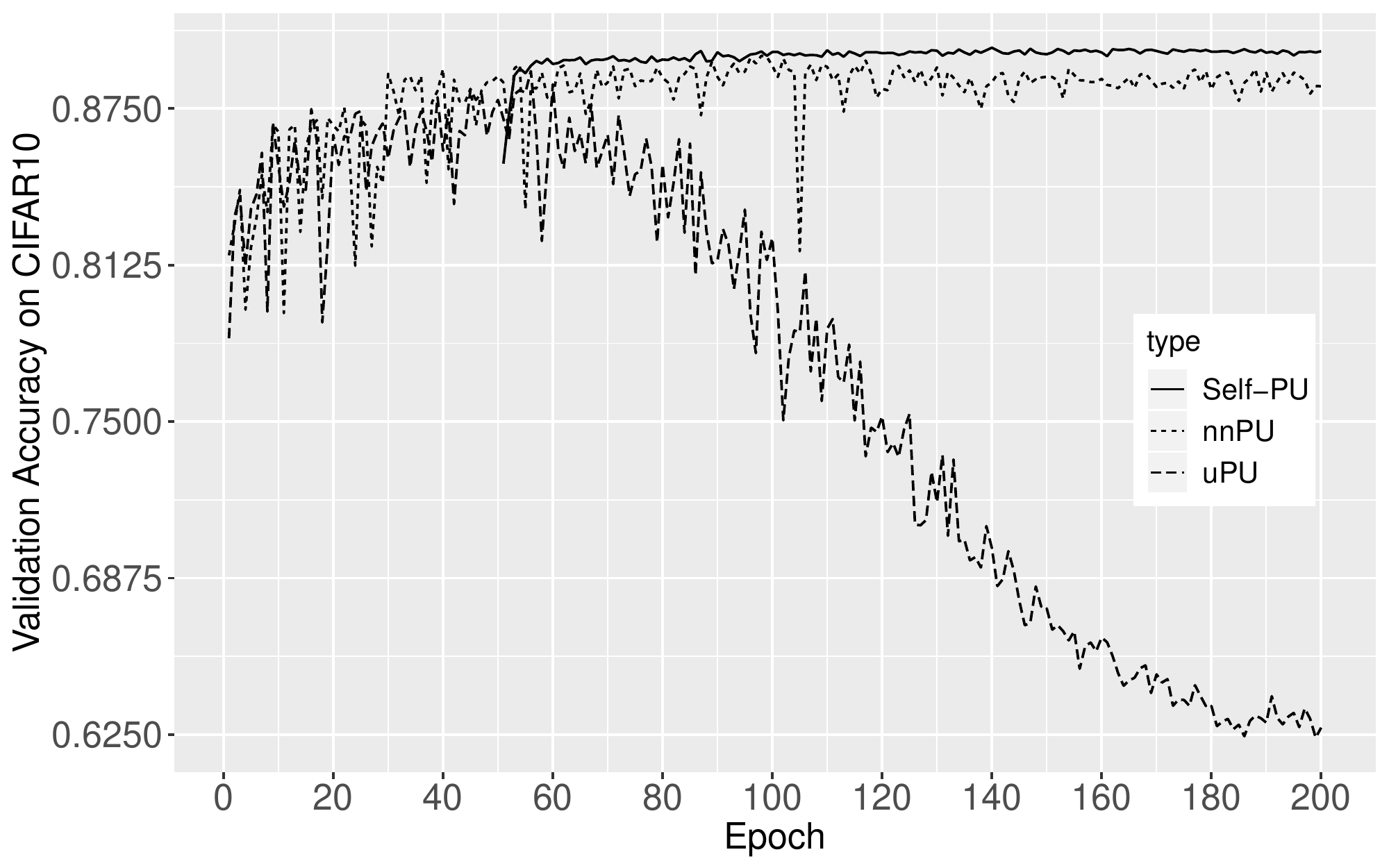}
\centering 
\caption{Validation accuracy during training on CIFAR-10. Our ``Self-PU'' framework\footnotemark achieved a more stable training compared with uPU and nnPU methods. }
\label{fig:training_validation_curve}
\end{figure}
\footnotetext{Since in ``Self-PU'' we use the teacher model $G$ for the final prediction, the solid line shows the accuracy of $G$ starts from epoch 50 when the self-paced training ends.}

Our ``Self-PU'' achieved not only high accuracy, but more importantly a \textbf{much more stable PU learning process} (Figure \ref{fig:training_validation_curve}). As noted in \cite{kiryo2017positive}, uPU suffers from overfitting complex models. We also empirically found a similar phenomenon in PU learning with nnPU risk estimator, where the validation accuracy remains unstable and even drops in the late training stage. However, the training process of Self-PU is significantly more stable than uPU and nnPU. This training stability benefits from both accurately identified examples in self-paced training and prediction consistency forced by our self-supervised distillations.

\subsubsection{Results on the new ADNI testbed}

Finally, we demonstrate the promise of our method on the more complex real-word ADNI data in Table \ref{table:sota_adni}. We first run a naive fully supervised classification baseline, by treating the entire \textbf{unlabeled} class as \textbf{negative}. Its accuracy is much inferior to our PU learning results, validating our PU formulation of the ADNI task. Next, Self-PU gains significantly over uPU and nnPU, showing highly promising performance on ADNI and setting new state-of-the-arts. Our sophisticated building blocks seem to add robustness to handling the real-world data variations and challenges. 

Furthermore, our above results seems to suggest that conventional PU benchmarks like CIFAR-10 and MNIST may have been saturated (as they already did in image classification). We recommend the community to pay more attention to more challenging and realistic new PU learning testbeds, and suggest ADNI as an effective, illustrative and practically important option. 

\section{Conclusion}
We proposed Self-PU, that bridges self-training strategy into PU learning for the first time. It leverages both the self-paced selected set of trusted samples and the consistency supervision via self-distillation and self-calibration. Experiments report state-of-the-art performance of Self-PU on two conventional (and potentially oversimplified) benchmarks, plus our newly introduced real-world PU testbed of ADNI classification. Our future work will explore more realistic PU learning setting, which we believe will motivate new algorithmic findings. 

\bibliography{main}
\bibliographystyle{icml2020}


\end{document}